\documentclass[10pt,twocolumn,letterpaper]{article}

\usepackage{iccv}
\usepackage{times}
\usepackage{epsfig}

\usepackage{amsmath}
\usepackage{amssymb}
\usepackage{booktabs}
\usepackage{adjustbox}
\usepackage{multirow}
\usepackage{bbding}
\usepackage{pifont}
\usepackage[table]{xcolor}
\usepackage[accsupp]{axessibility}

\newcommand \footnoteONLYtext[1]
{
	\let \mybackup \thefootnote
	\let \thefootnote \relax
	\footnotetext{#1}
	\let \thefootnote \mybackup
	\let \mybackup \imareallyundefinedcommand
}

\makeatletter\renewcommand\paragraph{\@startsection{paragraph}{4}{\z@}
  {.5em \@plus1ex \@minus.2ex}{-.1em}{\normalfont\normalsize\bfseries}}\makeatother

\usepackage[pagebackref=true,breaklinks=true,letterpaper=true,colorlinks,bookmarks=false]{hyperref}

\iccvfinalcopy 

\def\iccvPaperID{5628} 

\ificcvfinal\fi

\begin{document}

\title{Identity-Seeking Self-Supervised Representation Learning for \\ Generalizable Person Re-identification}

\author{Zhaopeng Dou$^{1,2}$, Zhongdao Wang$^{1,2}$, Yali Li$^{1,2}$, and Shengjin Wang$^{1,2*}$\\
$^1$ Department of Electronic Engineering, Tsinghua University, China\\
$^2$ Beijing National Research Center for Information Science and Technology (BNRist), China\\
{\tt\small dcp19@mails.tsinhua.edu.cn \quad\quad zhongdwang@gmail.com } \\
{\tt\small liyali13@tsinghua.edu.cn \quad\quad wgsgj@tsinghua.edu.cn$^*$}
}

\maketitle
\ificcvfinal\thispagestyle{empty}\fi
\footnoteONLYtext{$^*$ Corresponding author}

\begin{abstract}

This paper aims to learn a domain-generalizable (DG) person re-identification (ReID) representation from large-scale videos \textbf{without any annotation}. Prior DG ReID methods employ limited labeled data for training due to the high cost of annotation, which restricts further advances. To overcome the barriers of data and annotation, we propose to utilize large-scale unsupervised data for training. The key issue lies in how to mine identity information. To this end, we propose an Identity-seeking Self-supervised Representation learning (ISR) method. ISR constructs positive pairs from inter-frame images by modeling the instance association as a maximum-weight bipartite matching problem. A reliability-guided contrastive loss is further presented to suppress the adverse impact of noisy positive pairs, ensuring that reliable positive pairs dominate the learning process. The training cost of ISR scales approximately linearly with the data size, making it feasible to utilize large-scale data for training. The learned representation exhibits superior generalization ability. \textbf{Without human annotation and fine-tuning, ISR achieves 87.0\% Rank-1 on Market-1501 and 56.4\% Rank-1 on MSMT17}, outperforming the best supervised domain-generalizable method by 5.0\% and 19.5\%, respectively. In the pre-training$\rightarrow$fine-tuning scenario, ISR achieves state-of-the-art performance, with 88.4\% Rank-1 on MSMT17. The code is at \url{https://github.com/dcp15/ISR_ICCV2023_Oral}.

\end{abstract}

\section{Introduction}
\label{sec:introduction}
Person re-identification (ReID) aims to retrieve a person across non-overlapping camera views~\cite{PCB,ABD-Net,ISP,li2021combined,UAL}. Although current full-supervised ReID methods~\cite{Transreid,RFC,MoS,li2021diverse,wang2022pose,gu2022autoloss} have shown encouraging results on public benchmarks, their performance significantly declines when applied to unseen domains.
To enhance the generalization ability, unsupervised domain adaptation (UDA)~\cite{feng2021complementary,zhong2019invariance,fu2019self,bai2021hierarchical,rami2022online} and domain generalizable (DG) ReID~\cite{DIMN,jia2019frustratingly,RaMoE} techniques are extensively studied. However, both methods have limitations that prevent their scaling in real-world applications. 
UDA ReID necessitates well-organized data from the target domain for adaptation. DG ReID, on the other hand, typically employs small-scale labeled training data due to the high cost of annotation, which hinders further progress. To this end, we propose to break through the constraints of human annotation and target domain adaptation on the generalization of person ReID.

\begin{figure}[t]
    \centering
    \includegraphics[width=0.95\linewidth]{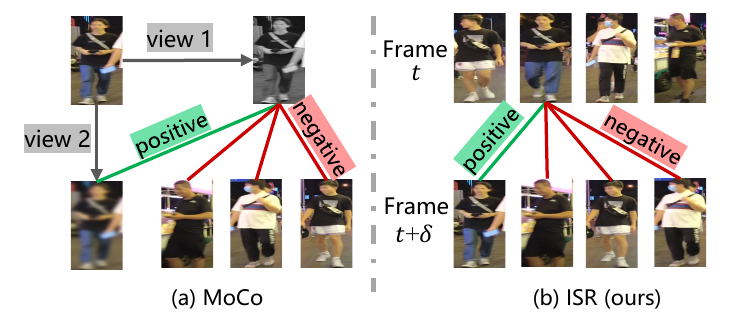}
    \caption{The core idea. (a) MoCo~\cite{MoCo} learns instance discrimination, which aims to learn a unique representation for each image. (b) ISR learns identity discrimination, which aims to learn similar representations for inter-frame images with the same identity.}
    \label{fig:introduction}
\end{figure}

This paper aims to learn domain-generalizable representations \textbf{\textit{without any annotation}}. The representation can be applied directly in arbitrary domains without fine-tuning. Unlike prior DG ReID methods~\cite{DIMN,jia2019frustratingly,RaMoE} that require labeled training data, we employ vast amounts of unlabeled internet videos for training. Specifically, our training set comprises 47.8 million unlabeled person images extracted from 74,000 video clips. The feasibility comes from two key factors: the low cost of acquiring videos from the internet and the diverse domains present in large-scale videos. We learn a domain-generalizable ReID representation from the large-scale unsupervised training data. The representation is robust to unseen domains, exhibiting high application potential and value for open-world scenarios.

The core issue is learning \textit{identity discrimination}. Inspired by the great success of contrastive learning~\cite{MoCo,BYOL,SimCLR,simsiam} in large-scale unsupervised data, we propose to learn identity discrimination through contrastive information between samples. It is not feasible to directly apply conventional unsupervised contrastive learning methods to ReID, because the objective of their pretext task (\ie, \textit{instance discrimination}) conflicts with ReID required identity discrimination. For example, as in Figure~\ref{fig:introduction}(a), MoCo~\cite{MoCo} regards two augmented views of an image as a positive pair, which prompts the model to learn a unique representation for each image. However, ReID aims to discriminate between identities instead of instances, which expects multiple images from the same identity to have similar representations. 

To achieve the objective of identity discrimination, we propose an Identity-seeking Self-supervised Representation learning (ISR) method. ISR aims to learn similar representations for inter-frame images belonging to the same identity, as illustrated in Figure~\ref{fig:introduction}(b). We first formulate the instance association between two frames as a maximum-weight bipartite matching problem, where positive pairs are mined by solving the optimal matching strategy. However, these positive pairs inevitably contain noise due to the unsupervised construction, which considerably undermines representation learning. Next, we introduce a reliability-guided contrastive loss to mitigate the adverse impact of noisy positive pairs, ensuring that reliable positive pairs dominate the learning process. Moreover, the training cost of ISR scales approximately linearly with the size of the data, making it feasible to train on large-scale data. 

Extensive experiments demonstrate the effectiveness of ISR. Under the domain-generalizable settings, ISR with ResNet50~\cite{resnet} backbone achieves 45.7\% Rank-1 and 21.2\% mAP on the most challenging dataset MSMT17~\cite{MSMT17}, outperforming the best-supervised domain-generalizable method trained with multiple labeled datasets by +8.8\% and +6.5\%, respectively. When using a more data-hungry backbone, \ie, Swin-Transformer~\cite{swin-transformer}, the performance is even further improved, achieving 56.4\% Rank-1 and 30.3\% mAP on MSMT17. We also evaluate ISR under other practical settings, such as pre-training for supervised fine-tuning or few-shot learning. ISR shows consistent improvements upon existing methods. For example, when serving as a pre-trained model for supervised fine-tuning, ISR achieves 88.4\% Rank-1 on MSMT17, setting a new state-of-the-art.

The main contributions of this paper are: 
\begin{itemize}
\item We propose an Identity-seeking Self-supervised Representation learning (ISR) method that can learn a domain-generalizable ReID representation from large-scale video data \textit{without any annotation}.

\item We propose a novel reliability-guided contrastive loss that effectively mitigates the adverse impact of noisy positive pairs, making reliable positive pairs dominate the representation learning process.

\item Extensive experiments verify the effectiveness of ISR. Notably, ISR achieves 87.0\% Rank-1 on Market-1501 and 56.4\% Rank-1 on MSMT17 without human annotation and fine-tuning, outperforming the best supervised DG method by 5.0\% and 19.5\%, respectively.
\end{itemize}

\section{Related Work}
\label{sec:related work}

\textbf{Person Re-identification.} Person re-identification (ReID)~\cite{PCB,ISP,UAL} aims to retrieve a person in a large database from disjoint cameras. Deep learning~\cite{Transreid,RFC,MoS,wang2022pose} has dominated the ReID community due to its powerful representation capabilities. The deep learning methods can be roughly divided into but not limited to, deep metric methods~\cite{tripletloss,hermans2017defense,circleloss}, part-based methods~\cite{PCB,MGN,UAL,guo2019beyond}, and attention-based methods~\cite{chen2019mixed,chen2019self,wang2022attentive}. Although these methods have achieved promising results on public benchmarks, they show poor generalization ability on unseen domains. As a result, unsupervised domain adaption (UDA) ReID ~\cite{rami2022online,bai2021hierarchical,zhong2019invariance} is proposed. UDA ReID aims at fine-tuning the source-trained model to the target domain using unlabeled target domain data. However, it is still not powerful enough for practical applications because it is sometimes difficult to collect well-organized target domain data for fine-tuning.
 
\textbf{Domain Generalizable (DG) ReID.} Domain generalizable ReID~\cite{DIMN,SNR,MetaBIN,M3L,RaMoE,MDA,DTIN-Net,DIR_ReID,ACL} aims to learn a robust model on the source domain and test it directly on unseen target domains without fine-tuning. It has attracted extensive attention due to its great potential in practical applications. DIMN~\cite{DIMN} designs a Domain-Invariant Mapping Network to learn domain-invariant representation under a meta-learning pipeline. MetaBIN~\cite{MetaBIN} and SNR~\cite{SNR} study the normalization layer or module to improve the model generalization. RaMoE~\cite{RaMoE} leverages the relevance between the target domain and multiple source domains to improve the model's generalization. MDA~\cite{MDA} aligns source and target feature distributions to a prior distribution. These methods are trained with small-scale domain-starved labeled data. Unlike them, we aim to learn a DG ReID model from large-scale domain-diverse unlabeled data.

\textbf{Synthetic Data for ReID.} 
The performance of ReID models is limited by the high cost of collecting annotated data from the real world. To address this challenge, several methods have turned to using synthetic data. Notably, PersonX~\cite{PersonX} contains 1,266 identities with 273,456 images captured from various viewpoints, enabling exploration of viewpoint impacts on ReID systems. RandPerson~\cite{RandPerson} offers 8,000 identities with 228,655 images from 19 cameras, while UnrealPerson~\cite{UnrealPerson} provides 3,000 identities with 120,000 images from 34 cameras, and ClonedPerson~\cite{ClonedPerson} includes 5,621 identities with 887,766 images from 24 cameras. These synthetic datasets have proven valuable for supervised learning, as they enhance the generalization ability of ReID models. DomainMix~\cite{Domain-mix} further establishes that combining labeled synthetic data with unlabeled real-world data during training is a promising direction for DG ReID. However, there remains a significant domain gap between synthetic and real-world data,  impeding the seamless application of models trained on synthetic data to authentic real-world scenarios. To bridge this disparity, we propose to use vast amounts of unlabeled real-world data for training.

\textbf{Unsupervised Representation Learning.} In recent years, unsupervised representation learning methods~\cite{MoCo, SimCLR, BYOL, MAE, simsiam} have shown promising results in learning pre-training representations from large-scale unlabeled data. MoCo~\cite{MoCo} introduces a memory queue to generate negative samples. SimCLR~\cite{SimCLR} leverages a projection head and rich data augmentations to achieve robust pre-training representation. BYOL~\cite{BYOL} learns the representation without using negative pairs. 
However, if these methods are applied directly to ReID, only a pre-training model can be learned, which shows extremely low accuracy when tested directly. The core reason is that they regard two views of an image as a positive pair or reconstruct masked pixels in an image, resulting in instance discrimination. This contradicts the ReID objective of identity discrimination. Unlike them, we regard inter-frame images with the same identity as positive pairs to align with the identity discrimination objective. A closely related work is CycAs~\cite{cycaseccv} and its improved version~\cite{CycAs}. CycAs enforces cycle consistency of instance association between video frame pairs to learn ReID representation. However, CycAs exhibits weaknesses in scenarios where certain instances lack corresponding matches or where association errors occur. Our method significantly differs from CycAs as we mine positive pairs and effectively suppress noise within them, offering a robust solution to learning a generalizable ReID representation.

\section{Identity-Seeking Represenatation Learning}
\label{sec:identity-preserving contrastive learning}

We present an Identity-seeking Self-supervised Representation (ISR) learning method to learn a domain-generalizable ReID representation from extensive videos without any annotation. ISR constructs positive pairs from inter-frame images to align with the ReID objective of \textit{identity discrimination} (Sec.~\ref{subsec:overview}). To mitigate the adverse effects of noisy positive pairs, we incorporate a reliability-guided contrastive loss (Sec.~\ref{subsec:noisy positive pair suppression}). The training cost of ISR scales approximately linear with the amount of training data, making it feasible to train with large-scale data.

\subsection{Overview}
\label{subsec:overview}

\begin{figure}[t]
    \centering
    \includegraphics[width=\linewidth]{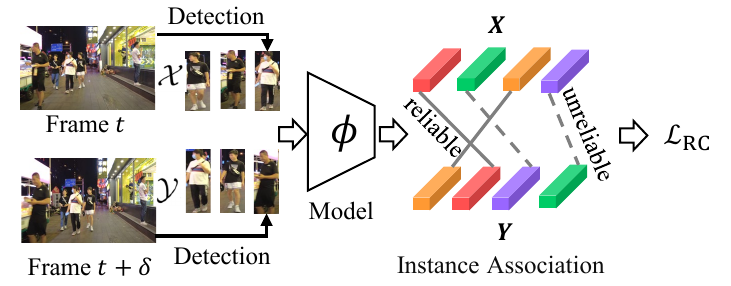}
    \caption{The overview of ISR. We first crop two image sets $\mathcal{X}$ and $\mathcal{Y}$ from a video frame pair. They are fed into the model $\phi$ to obtain the features $\boldsymbol{X}$ and $\boldsymbol{Y}$. We mine the positive pairs by modeling the association between $\boldsymbol{X}$ and $\boldsymbol{Y}$ as a maximum-weight bipartite matching problem. We propose a reliability-guided contrastive loss $\mathcal{L}_{\mathrm{RC}}$ to suppress the effect of noisy pseudo labels.}
    \label{fig:framework}
\end{figure}

The core of our method is to consider inter-frame images with the same identity as positive pairs for contrastive learning. As shown in Figure~\ref{fig:framework}, we crop the person images from two frames of a video with the help of an off-the-shelf pedestrian detection model~\cite{JDE} to form two image sets $\mathcal{X}$ and $\mathcal{Y}$. The time interval $\delta$ between two frames requires a trade-off. If $\delta$ is too large, $\mathcal{X}$ and $\mathcal{Y}$ might not have identity overlap. Conversely, if $\delta$ is too small, $\mathcal{X}$ and $\mathcal{Y}$ are too similar to provide sufficient contrastive information. In practice, we define a maximum time interval $\delta_{\mathrm{max}}$. We randomly select two frames as long as their interval does not exceed $\delta_{\mathrm{max}}$. $\mathcal{X}$ and $\mathcal{Y}$ are fed into a model $\phi$ to extract the features $\boldsymbol{X}=[\boldsymbol{x}_{1}, \boldsymbol{x}_{2},\dots, \boldsymbol{x}_{m}]\in\mathbb{R}^{d\times m}$ and $\boldsymbol{Y}=[\boldsymbol{y}_{1}, \boldsymbol{y}_{2},\dots, \boldsymbol{y}_{n}]\in\mathbb{R}^{d\times n}$, where $d$ is the feature dimension, $m$ and $n$ are the number of images in $\mathcal{X}$ and $\mathcal{Y}$, respectively. We assume $m\le n$; otherwise, we can swap $\boldsymbol{X}$ and $\boldsymbol{Y}$ to ensure this. 
All features are $l_2$-normalized.

The positive pairs are mined by formulating the instance association between $\boldsymbol{X}$ and $\boldsymbol{Y}$ as a maximum-weight bipartite matching problem. Let $\boldsymbol{\pi}\in \mathbb{R}^{m\times n}$ be a boolean association matrix, where $\boldsymbol{\pi}_{ij}=1$ if $\boldsymbol{x}_{i}$ and $\boldsymbol{y}_{j}$ are matched; otherwise, $\boldsymbol{\pi}_{ij}=0$. The cost of matching $\boldsymbol{x}_{i}$ and $\boldsymbol{y}_{j}$ is defined as the cosine distance between them, \ie, $c_{ij}=1 - \boldsymbol{x}_{i}\cdot\boldsymbol{y}_{j}$. One of our insights is that if the association matrix is optimal, the total matching cost should be minimum, which is formulated as:
\begin{equation}
\begin{aligned}
\min_{\boldsymbol{\pi}}  \quad  & \sum\nolimits_{ij}\boldsymbol{c}_{ij}\boldsymbol{\pi}_{ij}  \\
\mathrm{s.t.}  \quad & \sum\nolimits_{j}\boldsymbol{\pi}_{ij}=1, \quad &i\in \{1, 2, \dots, m\} \\
& \sum\nolimits_{i}\boldsymbol{\pi}_{ij}=0\ \mathrm{or}\ 1, \quad &j\in \{1,2,\dots, n\}
\end{aligned}
\label{eq:linear assignment}
\end{equation}
The optimal association matrix $\boldsymbol{\pi}^{*}$ can be solved in polynomial time by the Hungarian algorithm~\cite{hungarian} (please see supplementary materials for details). In general, there are finite pedestrians in a frame, so solving the optimal $\boldsymbol{\pi}^{*}$ brings limited computational overhead. This allows positive pairs to be mined synchronously during training, benefiting from each other with the learning of ReID model.

After solving the optimal association matrix $\boldsymbol{\pi}^{*}$, we define two matched samples as a positive pair. An intuitive way is to apply the loss function of MoCo~\cite{MoCo} to enforce the positive pairs to be similar and negative pairs to be dissimilar. However, such a way yields unsatisfactory results in our experiments (Table~\ref{table:ablation study of each components.}). We find such failure is caused by the inevitable noisy positive pairs. To address this issue, we propose a reliability-guided contrastive loss (Sec.~\ref{subsec:noisy positive pair suppression}) to suppress the impact of noisy positive pairs during training.

\paragraph{Identity discrimination \vs instance discrimination}. We analyze the feasibility of directly applying conventional unsupervised contrastive learning methods~\cite{MoCo, SimCLR, BYOL, simsiam} to ReID. Without loss of generality, we take the widely used MoCo~\cite{MoCo} as an example. MoCo considers two augmented views of an image to be a positive pair, which enforces the model to learn a unique representation for each image, leading to \textit{instance discrimination}. However, ReID necessitates \textit{identity discrimination}. Our method regards inter-frame images with the same identity as positive pairs and requires them to have similar representations, which aligns with the objective of ReID. As shown in Table~\ref{table:results of DG.}, under domain-generalizable settings, our method outperforms MoCo by a large margin, exhibiting great superiority in the ReID task.

\subsection{Noisy Positive Pair Suppression}
\label{subsec:noisy positive pair suppression}

There is inevitable noise in the mined positive pairs. Noisy positive pairs mainly come from two aspects: (1) the feature extractor $\phi$ is imperfect, which makes the matching cost inaccurate, especially in the early stages of training; (2) some pedestrians appearing in $\boldsymbol{X}$ do not appear in $\boldsymbol{Y}$, which is fatal because bipartite matching strategy constrains that for every sample in $\boldsymbol{X}$, there must be a sample in $\boldsymbol{Y}$ associated to it. 
Noisy positive pairs will seriously disrupt the learning of embedding space. To remedy this, we propose to measure the reliability of positive pairs and further suppress the adverse impact of noisy positive pairs during training.

\begin{figure}[t]
    \centering
    \includegraphics[width=\linewidth]{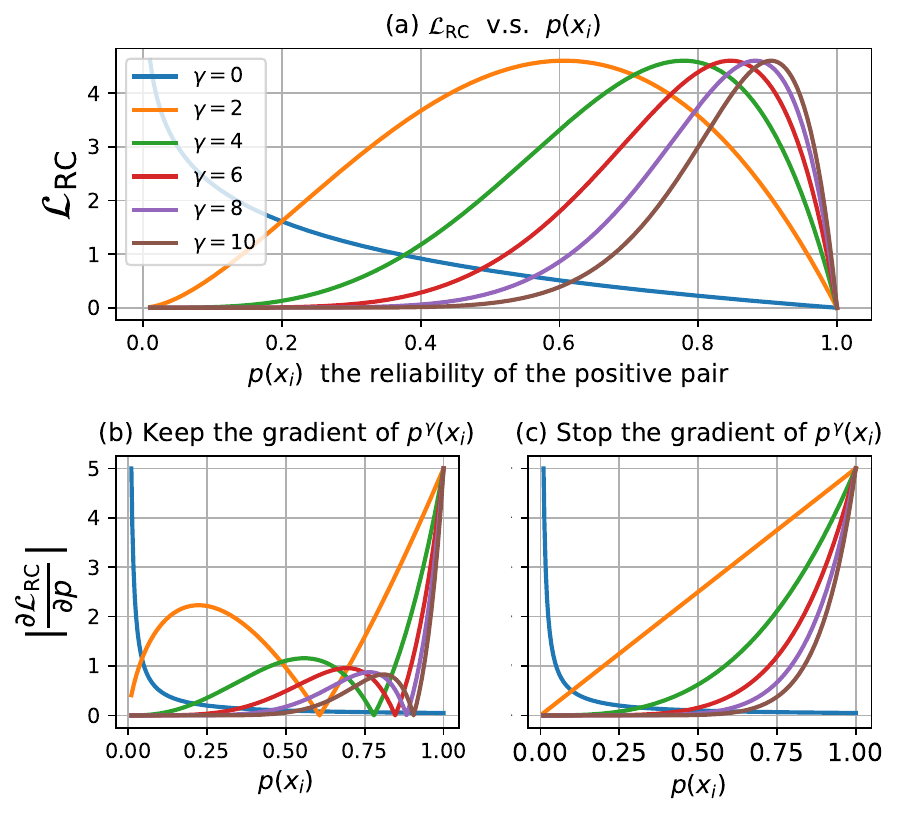}
    \caption{(a) The contrastive loss versus the reliability of the positive pair, \ie, $\mathcal{L}_{\mathrm{RC}}(\boldsymbol{x}_{i})$ \vs $p(\boldsymbol{x}_{i})$. Setting $\gamma>0$ suppresses the adverse impacts of positive pairs with low reliability scores. 
    (b) The absolute value of $\partial\mathcal{L}_{\mathrm{RC}}(\boldsymbol{x}_{i})/\partial p(\boldsymbol{x}_{i})$ when stopping the gradient of $p^{\gamma}(\boldsymbol{x}_{i})$.
    (c) The absolute value of $\partial\mathcal{L}_{\mathrm{RC}}(\boldsymbol{x}_{i})/\partial p(\boldsymbol{x}_{i})$ when keeping the gradient of $p^{\gamma}(\boldsymbol{x}_{i})$. 
    For a better observation, we scale up the curves under different $\gamma$ so that their maxima are equal.  }

    \label{fig:reliability-guided contrastive loss}
\end{figure}

Our intuition is that if two images belong to the same identity, their similarity should be large and vice versa. The similarity of two samples in the positive pair can reflect the reliability. If the similarity between two samples within a mined positive pair is very low, the positive pair is likely to be noisy. Besides, since the positive pairs are mined by solving the bipartite matching problem, the reliability should also consider the similarities between the target sample and other samples. Combing these two factors, for $\boldsymbol{x}_{i}$, the reliability score of the corresponding positive pair is,
\begin{equation}
    p(\boldsymbol{x}_{i}) = \frac{\sum\nolimits_{j}{\boldsymbol{\pi}^{*}_{ij}\exp(\boldsymbol{x}_{i}\cdot\boldsymbol{y}_{j}/\tau)}}{\sum\nolimits_{j}\exp(\boldsymbol{x}_{i}\cdot\boldsymbol{y}_{j}/\tau)}
    \label{eq:reliability score}
\end{equation}
where $\boldsymbol{\pi}^{*}$ is the optimal association matrix and $\tau$ is the temperature hyper-parameter. The reliability is the softmax-based probability that tries to align $\boldsymbol{x}_{i}$ with its positive one in the set $\boldsymbol{Y}$ . The calculation of reliability is extremely low overhead, making it suitable for large-scale training.

We propose the reliability score $p(\boldsymbol{x}_{i})$ to be a modulation factor to down-weight the noisy positive pairs and thus focus training on reliable positive pairs. At the same time, the modulation factor should not disrupt the gradient of the loss, so we break its gradient in back-propagation to ensure the stability of the training process. Formally, the reliability-guided contrastive loss is defined as,
\begin{equation}
    \mathcal{L}_{\mathrm{RC}}(\boldsymbol{x}_{i}) = -p^{\gamma}_{\nleftarrow}(\boldsymbol{x}_{i})\log(p(\boldsymbol{x}_{i}))
    \label{eq:reliability-guided contrastive loss}
\end{equation}
where $p^{\gamma}_{\nleftarrow}(\boldsymbol{x}_{i})$ indicates that the gradient of $p^{\gamma}(\boldsymbol{x}_{i})$ is cutoff in the back-propagation. $\gamma\ge0$ is the hyper-parameter controlling the strictness for reliability. The larger the $\gamma$, the more the loss function focuses on reliable positive pairs. In practice, the loss function $\mathcal{L}_{\mathrm{RC}}(\boldsymbol{x}_{i})$ will decrease as $\gamma$ increases. To maintain the size of the loss during training, we scale up the average $\mathcal{L}_{\mathrm{RC}}$ in the mini-batch,
\begin{equation}
    \mathcal{L}_{\mathrm{RC}} =  \frac{\alpha}{m}\sum\nolimits_{i=0}^{m} \mathcal{L}_{\mathrm{RC}}(\boldsymbol{x}_{i}) 
    \label{eq:scaled reliability-guided contrastive loss}
\end{equation}
where $\alpha=\left(\sum_{j}\mathcal{L}_{\mathrm{RC},\gamma=0}(\boldsymbol{x}_{j})\right)_{\nleftarrow} / \left( \sum_{j}\mathcal{L}_{\mathrm{RC}}(\boldsymbol{x}_{j})\right)_{\nleftarrow}$. This dynamically calibrates the size of the loss, weakening the influence of the learning rate in the optimal $\gamma$ analysis.

\paragraph{Properties of $\mathcal{L}_{\mathrm{RC}}(\boldsymbol{x}_{i})$}.  We visualize $\mathcal{L}_{\mathrm{RC}}(\boldsymbol{x}_{i})$ and the absolute gradient $\left| \partial\mathcal{L}_{\mathrm{RC}}(\boldsymbol{x}_{i})/\partial p(\boldsymbol{x}_{i}) \right|$ for $\gamma \in [0, 10]$ in Figure~\ref{fig:reliability-guided contrastive loss}. We note several key properties of $\mathcal{L}_{\mathrm{RC}}(\boldsymbol{x}_{i})$:

Firstly, as in Figure~\ref{fig:reliability-guided contrastive loss}a, when $\gamma=0$, $\mathcal{L}_{\mathrm{RC}}$ degenerates to a version without the reliability guidance. Noisy positive pairs comprise the majority of the loss, which seriously degrades representation learning. 
When $\gamma>0$, as shown in Figure~\ref{fig:reliability-guided contrastive loss}b, reliable positive pairs dominate the gradient back-propagation process, effectively mitigating the adverse impact of noisy positive pairs during learning.

Secondly, it is necessary to stop the gradient of $p^{\gamma}(\boldsymbol{x}_{i})$. If the gradient of $p^{\gamma}(\boldsymbol{x}_{i})$ is maintained, a slump point will occur for curves with $\gamma>0$, as in Figure~\ref{fig:reliability-guided contrastive loss}c. Consequently, positive pairs whose reliability lies in the trap lose the ability to optimize the model. Some noisy positive pairs exhibit large back-propagation gradients, severely compromising the representation learning. In our experiments, maintaining the gradient of $p^{\gamma}(\boldsymbol{x}_{i})$ will result in NAN issues. 

Lastly, $\mathcal{L}_{\mathrm{RC}}(\boldsymbol{x}_{i})$ significantly differs from the prior focal loss~\cite{focal_loss}. Focal loss is designed for training data with \textit{high-quality labels} by assigning large training weights to hard samples. $\mathcal{L}_{\mathrm{RC}}(\boldsymbol{x}_{i})$ is designed for training data with \textit{noisy pseudo-labels}, \ie, positive pairs with noise, by assigning large training weights to reliable positive pairs.
The modulating factor of focal loss is $(1-p)^{\gamma}$, while the modulating factor of $\mathcal{L}_{\mathrm{RC}}(\boldsymbol{x}_{i})$ is $p^{\gamma}_{\nleftarrow}$. The gradient cutoff of $p^{\gamma}$ is specific and crucial for $\mathcal{L}_{\mathrm{RC}}(\boldsymbol{x}_{i})$. Besides, results in Table~\ref{table:comparison with focal loss.} validate that $\mathcal{L}_{\mathrm{RC}}$ is superior to focal loss in our settings.

\subsection{Training and inference}
\label{subsec:training}
\paragraph{Training}. Inspired by~\cite{XBM,MoCo}, we adopt a memory queue to provide hard negative samples for contrastive learning. The queue uses a first-in, first-out update strategy. We regard two samples as a negative pair if they come from different videos. For $\boldsymbol{x}_{i}$, we select the most dissimilar $k$ negative samples from the queue, denoted as $\{\boldsymbol{f}_{1},\boldsymbol{f}_{2},\dots, \boldsymbol{f}_{k}\}$, to strengthen contrastive information, 
\begin{equation}
    \mathcal{L}_{\mathrm{Q}}(\boldsymbol{x}_{i})=\frac{1}{k}\sum\nolimits_{j=1}^{k}\log(1 + \exp{(\boldsymbol{x}_{i}\cdot \boldsymbol{f}_{j})})
    \label{eq:memory-based contrastive loss}
\end{equation}

Here, we introduce a ``super" frame sampling strategy to improve the efficiency of training. For video $b$, two cropped image sets are denoted as $\mathcal{X}_{b}$ and $\mathcal{Y}_{b}$. The ``super" frame pair is formed by merging several sets from different videos, \ie, $\mathcal{X}=\bigcup_{b}\mathcal{X}_{b}$ and $\mathcal{Y}=\bigcup_{b}\mathcal{Y}_{b}$. Note that the positive pairs are mined between two frames of the same video. The ``super" frame sampling strategy guarantees the domain diversity at each iteration.
Finally, the total loss function is:
\begin{equation}
\mathcal{L}=\mathcal{L}_{\mathrm{RC}} + \lambda \mathcal{L}_{\mathrm{Q}}
\label{eq:final loss}
\end{equation}
where $\lambda$ is a hyper-parameter to balance $\mathcal{L}_{\mathrm{RC}}$ and $\mathcal{L}_{\mathrm{Q}}$. 

\textit{Training cost analysis:} The number of pedestrians in a frame is limited, so the computational complexity of solving the bipartite-graph matching problem can be approximately regarded as $\mathcal{O}(1)$. Assuming that there are $N$ frames in our training data, the training complexity of our method is $\mathcal{O}(N)$, which is roughly linear with the data size. This makes it possible to drive large-scale data for training. Please refer to the supplementary for related experiments. 

\paragraph{Inference}. If not specified, the learned representation is directly tested on target domains without fine-tuning.

\section{Experiments}
\label{sec:experiments}

\subsection{Implementation}
\label{subsec:implementation}

\begin{table*}[t]
\small
\begin{center}
\adjustbox{max width=1.0\linewidth}{
\begin{tabular}{l|c|cc|cc|cc|cc|cc|cc|cc}
\toprule
\multirow{3}{*}{Methods}  & \multirow{3}{*}{Sup.} & \multicolumn{6}{c|}{Protocol-1} & \multicolumn{8}{c}{Protocol-2} \\ \cline{3-16}  
          &       &  \multicolumn{2}{c|}{Market-1501}  &  \multicolumn{2}{c|}{MSMT17}  &  \multicolumn{2}{c|}{CUHK03}  &  \multicolumn{2}{c|}{PRID}   &  \multicolumn{2}{c|}{GRID} &  \multicolumn{2}{c|}{VIPeR}     &  \multicolumn{2}{c}{iLIDs}   \\  \cline{3-16}
          &       & R1     & mAP         & R1   & mAP    & R1   & mAP    & R1     & mAP         & R1   & mAP     & R1     & mAP  & R1   & mAP \\ \hline
CrossGrad*~\cite{CrossGrad}  & \ding{52}  &  --  &  --  & -- & -- & -- & -- & 18.8 & 28.2  & 9.0   & 16.0  &  20.9  & 30.4  & 49.7 & 61.3  \\
MLDG*~\cite{MLDG}            & \ding{52}  &  --  &  --  & -- & -- & -- & -- & 24.0 & 35.4  & 15.8  & 23.6  &  23.5  & 33.5  & 53.8 & 65.2  \\
PPA*~\cite{PPA}              & \ding{52}  &  --  &  --  & -- & -- & -- & -- & 31.9 & 45.3  & 26.9  & 38.0  &  45.1  & 54.5  & 64.5 & 72.7  \\  
DIMN*~\cite{DIMN}            & \ding{52}  &  --  &  --  & -- & -- & -- & -- & 39.2 & 52.0  & 29.3  & 41.1  &  51.2  & 60.1  & 70.2 & 78.4  \\ 
SNR~\cite{SNR}               & \ding{52}  &  --  &  --  & -- & -- & -- & -- & 52.1 & 66.5  & 40.2  & 47.7  &  52.9  & 61.3  & 84.1 & 89.9  \\
ACL~\cite{ACL}     & \ding{52}  &  --  & --  & -- & -- & -- & -- & 63.0 & 73.4 & 55.2 & 65.7 & \textbf{66.4} & \textbf{75.1} & 81.8 & 86.5 \\
MetaBIN~\cite{MetaBIN}       & \ding{52}  &  --  &  --  & -- & -- & -- & -- & \textbf{74.2} & \textbf{81.0}  & 48.4  & 57.9  &  59.9  & 68.6  & 81.3 & 87.0  \\ 
MDA~\cite{MDA}               & \ding{52}  &  --  &  --  & -- & -- & -- & -- & -- & --    & \textbf{61.2}  & \textbf{62.9}  &  63.5  & 71.7  & 80.4 & 84.4  \\ 
DDAN~\cite{DDAN}             & \ding{52}  &  --  &  --  & -- & -- & -- & -- & 62.9 & 67.5  & 46.2  & 50.9  &  56.5  & 60.8  & 78.0 & 81.2  \\
DTIN~\cite{DTIN-Net}         & \ding{52}  &  --  &  --  & -- & -- & -- & -- & 71.0 & 79.7  & 51.8 & 60.6   & 62.9   & 70.7 & 81.8 & 87.2 \\ 
META~\cite{META}             & \ding{52}  &  --  &  --  & -- & -- & -- & -- & 61.9 & 71.7  & 51.0 & 56.6   & 53.9   & 60.4 & 79.3 & 83.9 \\
M$^3$L~\cite{M3L}            & \ding{52} & 75.9 & 50.2 & \textbf{36.9} & \textbf{14.7}  & 33.1 & 32.1  &  --  &  --  & -- & -- & -- & -- & -- & --        \\
 DML~\cite{RaMoE}             & \ding{52}   & 75.4 & 49.9 & 24.5 & 9.9 & 32.9 & 32.6  & 47.3 & 60.4 & 39.4 & 49.0  & 49.2  & 58.0 & 77.3  & 84.0 \\
RaMoE~\cite{RaMoE}           & \ding{52}  & \textbf{82.0} & \textbf{56.5} & 34.1 & 13.5  & \textbf{36.6} & \textbf{35.5} & 57.7 & 67.3 & 46.8 & 54.2  & 56.6  & 64.6 & \textbf{85.0}  & \textbf{90.2} \\  \hline
TrackContrast~\cite{TrackContrast}  & \ding{56} & 72.7 & 36.2    & -- & -- & -- & --  &  --  &  --  & -- & -- & -- & -- & -- & --  \\  
LUP$^\dagger$~\cite{LUP}               &  \ding{56} & 3.3  & 1.0   & 0.3  & 0.1   & 0.1   & 0.5  & 1.5  & 3.7  & 1.2 & 4.0    & 1.4    & 5.0 & 36.7     & 43.0   \\
MoCo (R50)~\cite{MoCo}  &  \ding{56} & 10.5 & 2.6 & 0.5 & 0.2 & 0.3 & 0.7 & 6.5 & 10.9 & 2.8 & 6.9 & 4.0 & 7.5 & 38.8 & 46.4 \\
LUPnl$^\dagger$~\cite{LUP-NL}          & \ding{56} & 13.8 & 3.8  & 0.6  & 0.2     & 0.4   & 0.8  & 8.1   & 12.2   & 3.1   & 7.4    & 4.6 & 9.2   & 43.3    & 49.8 \\  
CycAs (R50)~\cite{CycAs} & \ding{56} & 80.3  & 57.5  &  43.9  & 20.2  & 25.8  & 26.5   & 58.8 & 67.7  & 52.5  & 62.3  & 57.3  & 66.0  & 85.2  & 90.4 \\   
\textbf{Ours (R50)}               &  \ding{56} & \textbf{85.1}   & \textbf{65.1} & \textbf{45.7}   & \textbf{21.2}       & \textbf{26.1}   & \textbf{27.4}     & \textbf{59.7}   & \textbf{70.8}   & \textbf{55.8}   & \textbf{65.2}  & \textbf{58.0}  & \textbf{66.6}   & \textbf{87.6}    & \textbf{91.7}   \\  \hline
CycAs (Swin)~\cite{CycAs} & \ding{56}  & 82.2  & 60.4 &  49.0  & 24.1   & 36.3  & 37.1  & 71.5 & 79.2  & 55.8  & 66.4  & 60.2  & 68.8  & 87.4  & 91.3 \\ 
\textbf{Ours (Swin)}                    &  \ding{56}  & \textbf{87.0}   & \textbf{70.5} & \textbf{56.4}    & \textbf{30.3}     & \textbf{36.6}   & \textbf{37.8}    &  \textbf{74.5}  & \textbf{83.0}   & \textbf{62.7}   & \textbf{72.0}  & \textbf{68.4}  & \textbf{75.5} & \textbf{87.5} & \textbf{91.5}   \\

\bottomrule
\end{tabular}
}
\end{center}
\caption{ 
Comparison with state-of-the-art under DG settings. The performances of the methods marked by ``*" are from \cite{DIMN}. ``Sup." indicates the method is supervised or unsupervised. $\dagger$ indicates the results are tested with the pre-trained models published on their Github.
}

\label{table:results of DG.}
\end{table*}

\begin{table*}[t]
\small
    \centering
    \adjustbox{max width=\linewidth}{
    \begin{tabular}{c|c|c|cc|cc|cc|cc}
    \toprule
         \multirow{2}{*}{Method} & \multirow{2}{*}{Sup.} & \multirow{2}{*}{Training set}  & \multicolumn{2}{c|}{PersonX$_{456}$}  & \multicolumn{2}{c|}{UnrealPerson} & \multicolumn{2}{c|}{ClonedPerson}   & \multicolumn{2}{c}{RandPerson} \\ 
         &  &   &  R1 & mAP & R1 & mAP & R1 & mAP & R1 & mAP \\ \hline
         BOT~\cite{BOT} & \ding{52} & M+D+C3+MT  & 87.7 & 72.7 & 56.1 & 43.5 & 41.0 & 7.9 & 35.8 & 16.8  \\ 
         MGN~\cite{MGN} & \ding{52}  & M+D+C3+MT  & 91.2    & 77.8  & 58.1 & 45.7 & 49.1 & 11.9 & 50.1 & 27.1     \\ \hline
         MoCo (R50)~\cite{MoCo} & \ding{56}  & Unsup-videos   & 16.0  & 1.4  & 0.7 & 0.2 & 0.5 & 0.2 & 0.2 & 0.1  \\
         \textbf{Ours (R50)} & \ding{56}  & Unsup-videos    & \textbf{92.5}  & \textbf{81.6} & \textbf{64.6}  & \textbf{52.8} & \textbf{49.8} & \textbf{12.1}  & \textbf{54.2} & \textbf{28.1}  \\
         \textbf{Ours (Swin)} & \ding{56}  & Unsup-videos    &  \textbf{95.0}  & \textbf{88.5} & \textbf{67.3} &  \textbf{58.1}  & \textbf{57.6} & \textbf{16.6} & \textbf{68.3} & \textbf{40.1} \\
    \bottomrule
    \end{tabular}
    }
    \caption{Comparison on the synthetic datasets: PersonX$_{456}$~\cite{PersonX}, UnrealPerson~\cite{UnrealPerson}, ClonedPerson~\cite{ClonedPerson} and RandPerson~\cite{RandPerson}. We use UnrealPerson-v1.3 and the subset of RandPerson for testing, each  partitioned into query and gallery sets with a ratio of 1:4.
    ``unsup-videos" indicates our large-scale unlabeled video dataset. M: Market~\cite{market}, D: Duke~\cite{duke}, C3: CUHK03~\cite{CUHK03}, MT: MSMT17~\cite{MSMT17}.  }
    \label{table:results of personX}
\end{table*}

\begin{table*}[t]
\begin{center}
\adjustbox{max width=\linewidth}{
\setlength{\tabcolsep}{3.0pt}
\begin{tabular}{c|c|ccccc|ccccc}
\toprule
 \multirow{2}{*}{Datasets}  &   \multirow{2}{*}{Pre-train}  &  \multicolumn{5}{c|}{Small-scale}  &  \multicolumn{5}{c}{Few-shot} \\ \cline{3-12}
  &   &  10\%  & 30\%  & 50\%  & 70\%  & 90\%  &  10\%  & 30\%  & 50\%  & 70\%  & 90\%    \\   \hline
\multirow{5}{*}{Market} & IN sup. & 76.9 / 53.1  &  90.8 / 75.2  & 93.5 / 81.5  & 94.5 / 84.8  & 95.2 / 86.9 & 41.8 / 21.1 & 87.6 / 68.1 & 92.8 / 80.2 & 94.0 / 84.2 & 94.6 / 86.7\\
                        & IN unsup. & 81.7 / 58.4  &  91.9 / 76.6  & 94.1 / 82.0  & 94.5 / 85.4  & 95.5 / 87.4 & 36.1 / 18.6 & 87.8 / 69.3 & 90.9 / 78.3 & 94.1 / 84.4 & 95.2 / 87.1 \\
                        & MoCo~\cite{MoCo}  &  81.8 / 58.8 &  92.3 / 77.7 & 94.3 / 84.0 &  95.4 / 87.3 & 95.9 / 89.2    & 41.5 / 22.0 & 87.7 / 69.8  & 93.9 / 83.1 &  94.8 / 86.8  &  96.0 / 89.3  \\
                        & LUP~\cite{LUP} & 85.5 / 64.6  &  93.7 / 81.9  & 94.9 / 85.8  & 95.9 / 88.8  & 96.4 / 90.5 & 47.5 / 26.4 & 92.1 / 78.3 & 93.9 / 84.2 & 95.5 / 88.4 & 96.3 / 90.4 \\
                        & LUPnl~\cite{LUP-NL} & 88.8 / 72.4  & 94.2 / 85.2  & 95.5 / 88.3  & 96.2 / 90.1 & 96.4 / 91.3  & 61.6 / 42.0  & 94.0 / 83.7  & 95.2 / 88.1  & 96.3 / 90.5  & 96.4 / 91.6  \\ \cline{2-12}
                        & \textbf{Ours} & \textbf{90.5 / 75.3}  & \textbf{94.8 / 86.2}  & \textbf{96.2 / 89.3}  & \textbf{96.6 / 90.9} & \textbf{96.7 / 91.8}  & \textbf{75.9 / 54.3}  & \textbf{94.1 / 84.9}  & \textbf{96.0 / 89.4}  & \textbf{96.4 / 91.3}  & \textbf{96.8 / 92.1}  \\ \hline

\multirow{5}{*}{MSMT17} & IN sup. & 50.2 / 23.2  &  70.8 / 41.9  & 76.9 / 50.3  & 81.2 / 56.9  & 84.2 / 61.9 & 34.1 / 14.7 & 71.1 / 44.5 & 79.5 / 56.2 & 82.8 / 60.9 & 84.5 / 63.4 \\
                        & IN unsup. & 48.8 / 22.6  &  68.7 / 40.4  & 75.0 / 49.0  & 79.9 / 55.7  & 83.0 / 60.9 & 29.2 / 13.2 & 67.1 / 41.4 & 77.6 / 53.3 & 81.5 / 59.1 & 83.8 / 62.4 \\
                        & MoCo~\cite{MoCo} & 46.8 / 21.5 & 69.4 / 41.7 & 76.7 / 50.8  & 81.3 / 58.3  & 84.7 / 63.9   & 23.4 / 9.9   &  65.9 / 40.7   &  77.4 / 54.2   & 82.7 / 61.5     &  85.0 / 65.3 \\    
                        & LUP~\cite{LUP} & 51.1 / 25.5  &  71.4 / 44.6  & 77.7 / 53.0  & 81.8 / 59.5  & 85.0 / 63.7 & 36.0 / 17.0 & 73.6 / 49.0 & 80.5 / 57.4 & 83.5 / 62.9 & 85.1 / 65.0 \\
                        & LUPnl~\cite{LUP-NL} & 51.1 / 28.2  & 71.2 / 47.7  & 77.2 / 55.5  & 81.8 / 61.6 & 84.8 / 66.1  & 42.7 / 24.5  & 74.4 / 53.2  & 81.0 / 62.2  & 83.8 / 65.8  & 85.3 / 67.4  \\  \cline{2-12}
                        & \textbf{Ours} & \textbf{64.2 / 36.2}  & \textbf{78.1 / 52.9}  & \textbf{82.5 / 60.0}  & \textbf{85.4 / 65.7} & \textbf{87.5 / 69.6}  & \textbf{59.4 / 33.4}  & \textbf{80.9 / 59.3}  & \textbf{85.6 / 66.3}  & \textbf{87.6 / 69.7}  & \textbf{88.1 / 70.9}  \\ 
                        
\bottomrule
\end{tabular}
}
\end{center}
\caption{Comparisons under the \textit{small-scale} settings and \textit{few-shot} settings. Results are shown as \textit{Rank-1 / mAP}.}
\label{table:results of small-scale and few-shot.}
\end{table*}

\textbf{Training dataset.} Following LUP~\cite{LUP}, LUPnl~\cite{LUP-NL}, and CycAs~\cite{CycAs}, we crawl the videos from youtube as our training data. The videos are searched by queries like ``cityname+streeview". The ``cityname" includes 100 big cities around the world, ensuring the diversity of the domains. Following~\cite{CycAs},
to build a high-quality training dataset, the processing includes: (1) use the PySceneDetect\footnote{\url{http://scenedetect.com/}} toolkit to cut the raw video into multiple slices, eliminating the effects of shot changes within a video; (2) use the off-of-shelf pedestrian detection model JDE~\cite{JDE} to crop the person images every seven frames to reduce the temporal redundancy; (3) calculate the average number of pedestrians detected in five uniformly sampled frames in a video, removing videos with less than five pedestrians. Statistically, the training set contains 47.8M person images from 74K video clips.

\textbf{Test datasets and protocols.} We conduct extensive experiments on public ReID datasets. For domain generalizable (DG) ReID, following~\cite{RaMoE}, we select two common protocols. Protocol-1 is the leave-one-out setting for Market-1501\cite{market}, DukeMTMC~\cite{duke}, CUHK03~\cite{CUHK03}, and MSMT17~\cite{MSMT17}, which selects one domain for testing (only the testing set) and all the other domains for training (including the training and testing sets). Protocol-2 selects all images in Market-1501, DukeMTMC, CUHK02~\cite{CUHK02}, CUHK03, and CUHK-SYSU~\cite{CUHK-SYSU} (including the training and testing sets) as the training images. Four small datasets, \ie, PRID~\cite{PRID}, GRID~\cite{grid}, VIPeR~\cite{VIPeR}, and iLIDs~\cite{iLIDs} are used for testing. Following~\cite{DIMN}, we report the average of 10 random splits of gallery and probe sets. \textbf{\textit{Since DukeMTMC has been withdrawn, we do not report the evaluation results on DukeMTMC}}, which is introduced just to illustrate the training settings of other methods.

\textbf{Training details.} If not specified, we use ResNet50~\cite{resnet} as our backbone. Swin-Transformer~\cite{swin-transformer} is also adopted for fair comparisons.
Similar to~\cite{BOT}, the stride of ResNet50's last layer is set to 1.
Inspired by~\cite{IBN}, we add one IN~\cite{IN} layer to the end of layer1 and layer2 of ResNet50, respectively, to improve the model generalization. 
For ResNet50 (Swin-Transformer), input images are resized to $256\times128$ ($224\times224$). During training, images are augmented by random horizontal flipping and color jitter. We find suitable data augmentation is beneficial for unsupervised contrastive learning. Please refer to supplementary materials for related ablation studies. To guarantee the stability of training, the number of images in a "super" frame is 80 (more than 80 will be truncated) and we sample three frames from each video and team each other up to form three frame pairs. We adopt distributed training on $4\times$NVIDIA 3090GPU, so the batch size is $960=80\times3\times4$ (for Swin-Transformer, it is $336=28\times3\times4$). In an epoch, we sample every video 16 times. The model is trained for 50 epochs with the AdamW~\cite{AdamW}. We use the Cosine Annealing learning rate~\cite{CosineAnnealing} initialized with 1$e^{-4}$. We set the hyper-parameter $\gamma=6$ and $\lambda=5$ in Eq.~\ref{eq:reliability-guided contrastive loss} and Eq.~\ref{eq:final loss}, respectively.

\textbf{Evaluation metrics:} Cumulative matching characteristic (CMC) curve and mean average precision (mAP).

\subsection{Domain Generalization (DG) Setting.}
\label{subsec:domain generalizable ReID}
There are few unsupervised methods for domain generalizable (DG) ReID. Therefore we mainly compare our method with supervised DG ReID methods. We also compare five unsupervised methods that utilize large-scale video data similar to ours. \textbf{\textit{Notably, our method uses the crawled unsupervised videos for training, and tests directly on these datasets.}} We report the results in Table~\ref{table:results of DG.}. Results on DukeMTMC are not reported as it is retracted. 

\textit{Compared to supervised methods:}
(1) When adopting ResNet50 as the backbone, the state-of-the-art supervised method to be compared is RaMoE~\cite{RaMoE}. Compared with RaMoE, our method with ResNet50 backbone outperforms it on most datasets (six out of seven). Especially in the most challenging and largest dataset MSMT17, our method outperforms it by 11.6\% Rank-1 and 7.7\% mAP.
(2) When adopting a more advanced network, \ie, Swin-Transformer, as the backbone, the performance of our method is further improved and outperforms all considered supervised DG methods on all listed datasets without any human annotation, refreshing state-of-the-art. This demonstrates that data-hungry networks can make better use of large-scale training data, which is more suitable for our method.
(3) These results show the paradigm of combining unsupervised contrastive learning and domain-diverse large-scale unlabeled training data is indeed one of the feasible directions for DG ReID, which can learn a DG ReID model even better than supervised methods with limited training data.

\textit{Compared to unsupervised methods:} we also compare our method with five unsupervised methods, \ie, TrackContrast~\cite{TrackContrast}, LUP~\cite{LUP}, LUPnl~\cite{LUP-NL}, MoCo~\cite{MoCo} and CycAs~\cite{CycAs}. 
Our method outperforms TrackContrast by a large margin. For example, our method with ResNet50 backbone outperforms it by 28.9\% mAP on Market-1501. Models trained by LUP and LUPnl can only serve as pre-training models and perform poorly when used for \textit{direct testing}. The results of LUP\footnote{\url{https://github.com/DengpanFu/LUPerson}} and LUPnl\footnote{\url{https://github.com/DengpanFu/LUPerson-NL}} are tested by us using their public models. CycAs uses the same data scale as ours. With the ResNet50 backbone, our method outperforms CycAs by 7.6\% mAP in Market-1501. With the Swin-transformer backbone, our method suppresses CycAs by 10.1\% mAP in Market-1501.
The comparisons between our method and MoCo are fair because the MoCo is trained with our crawled video data. Our method outperforms MoCo significantly on all listed datasets. This demonstrates that the instance discrimination learned by MoCo and the identity discrimination required by ReID are misaligned. Our method focuses on learning identity-discriminative representations, which is consistent with the objective of ReID.

\begin{table}[t]
\small
\begin{center}
\adjustbox{max width=\linewidth}{
\setlength{\tabcolsep}{5pt}
\begin{tabular}{l|cc|cc|cc}
\toprule
\multirow{2}{*}{Methods}  & \multicolumn{2}{c|}{CUHK03} & \multicolumn{2}{c|}{Market-1501} & \multicolumn{2}{c}{MSMT17} \\  \cline{2-7}
                       & R1     & mAP         & R1   & mAP      & R1   & mAP \\ \hline
PCB~\cite{PCB}                 & 63.7   & 57.5    & 93.8    &  81.6         &  --    &  --     \\
MGN$\dagger$~\cite{MGN}        & 71.2   & 70.5    & 95.1    &  87.5         &  85.1  &  63.7    \\
BOT~\cite{BOT}                 &  --    &  --     & 94.5    &  85.9         &  --    &  --      \\
DSA~\cite{DSA}                 & 78.9   & 75.2    & 95.7    &  87.6         &  --    &  --      \\
ABDNet~\cite{ABDNet}           & --     & --      & 95.6    &  88.3         & 82.3   &  60.8    \\
OSNet~\cite{OSNet}             & 72.3   & 67.8    & 94.8    &  84.9         & 78.7   &  52.9    \\   
MHN~\cite{MHN}                 & 77.2   & 72.4    & 95.1    &  85.0         &  --    &  --      \\
BDB~\cite{BDB}                 & 79.4   & 76.7    & 95.3    &  86.7         &  --    &  --      \\
GCP~\cite{GCP}                 & 77.9   & 75.6    & 95.2    &  88.9         &  --    &  --      \\
ISP~\cite{ISP}                 & 76.5   & 74.1    & 95.3    &  88.6         &  --    &  --      \\
GASM~\cite{GASM}               &  --    & --      & 95.3    &  84.7         & 79.5   &  52.5    \\
ESNET~\cite{ESNET}             &  --    & --      & 95.7    & 88.6          & 80.5   &  57.3    \\  \hline
*IN sup.+MGN$\dagger$                  & 71.2   & 70.5    & 95.1    & 87.5      & 85.1   & 63.7     \\
*IN unsup.+MGN$\dagger$                & 67.0   & 67.1    &  95.3     & 88.2       & 84.3   & 62.7     \\
MoCo~\cite{MoCo}+MGN$\dagger$  & 73.2 & 71.8 & 96.2 & 90.3 & 85.2 & 65.5 \\ 
LUP~\cite{LUP}+MGN$\dagger$             & 75.4   & 74.7    & 96.4    & 91.0          & 85.5   &  65.7     \\                     
CycAs~\cite{CycAs}+MGN$\dagger$      & 76.9   & 76.3   & 96.5  & 91.2    & 86.1   & 65.8   \\ 
LUPnl~\cite{LUP-NL}+MGN$\dagger$     & 80.9   & 80.4    & 96.6    & 91.9   & 86.0  & 68.0      \\  \hline
\textbf{Ours+MGN$\dagger$}                  & \textbf{82.7}  & \textbf{81.7}    & \textbf{96.9}    & \textbf{92.3}          & \textbf{88.4}   & \textbf{71.5}      \\
\bottomrule
\end{tabular}
}
\end{center}
\caption{Comparisons under the pre-training $\rightarrow$ fine-tuning settings. Results of methods marked by ``*" are from~\cite{LUP-NL}. ``IN sup."/``IN unsup." indicates model that is pretrained on ImageNet in a supervised/unsupervised manner. MGN$\dagger$ refers to the re-implementation of MGN in fast-reid.}
\label{table:results of fine-tuning.}
\end{table}

\subsection{DG Setting on Synthetic Dataset}
\label{subsec:DG Setting on Synthetic Dataset}
We further conduct experiments on the synthetic datasets, including PersonX$_{456}$~\cite{PersonX}, UnrealPerson~\cite{UnrealPerson}, ClonedPerson~\cite{ClonedPerson} and RandPerson~\cite{RandPerson}. Since there is almost no method to conduct DG experiments on these synthetic datasets, we compare our method with two strong supervised baselines BOT~\cite{BOT} and MGN~\cite{MGN}, and one unsupervised method MoCo~\cite{MoCo}. Both BOT and MGN use the combination of Market-1501, Duke, CUHK03, and MSMT17 for supervised training. Our method and MoCo are trained with the crawled large-scale unlabeled video data and directly tested on these synthetic datasets without fine-tuning. Results are reported in Table~\ref{table:results of personX}. Our method outperforms BOT, MGN and MoCo by a large margin, especially in terms of mAP. This shows that our method generalizes well, even if the target and source domains are essentially different. Our method learns the core concepts of pedestrian representation in an unsupervised manner.

\subsection{Small-scale and Few-shot Settings.}
\label{subsec:comparison on small-scale and few-shot}
Following LUP~\cite{LUP} and LUPnl~\cite{LUP-NL}, we also conduct experiments under two small data settings: the \textit{small-scale} setting and the \textit{few-shot} setting. The small-scale setting controls the percentage of usable IDs, and the few-shot setting controls the percentage of images for each ID. These two settings can reflect the adaptability of the model in new scenarios. We vary the usable percentages from 10\% to 90\% and use the MGN implemented in fast-reid as the baseline. Results are reported in Table~\ref{table:results of small-scale and few-shot.}. Our method comprehensively outperforms all other pre-trained models. This shows our model can adapt to new environments more efficiently and has a more substantial potential for real-world applications. 
\textit{Note} that under the few-shot setting with 10\% images per identity for fine-tuning, on Market-1501, the performance is lower than direct evaluation, \ie, 75.9/54.3 \vs 85.1/65.1 (in Table~\ref{table:results of DG.}). This is because when the usable images for each identity are few, it is insufficient to train the classifier with multiple classes well, which in turn compromises the feature extractor. Therefore, when the data for fine-tuning is noisy or poorly organized, we recommend using our model for direct testing without fine-tuning, which further demonstrates the superiority of our method.

\subsection{Pre-training$\rightarrow$Fine-tuning Setting.}
\label{subsec:serve as pretrained models}
Following~\cite{LUP} and ~\cite{LUP-NL}, we evaluate the effectiveness of the model learned by our method as a pre-trained model. We conduct experiments on the strong baseline MGN~\cite{MGN} implemented by fast-reid with six different pre-training models, \ie, ``IN sup.", ``IN unsup.", LUP~\cite{LUP}, LUPnl~\cite{LUP-NL}, MoCo~\cite{MoCo} and CycAs~\cite{CycAs}. ``IN sup."/``IN unsup." indicates the model is pre-trained on ImageNet in a supervised/unsupervised manner. MoCo is pre-trained on our collected data. 
The results are shown in Table~\ref{table:results of fine-tuning.}. Note that post-processing such as IIA~\cite{IIA} and RR~\cite{ReRank} are not applied. We can see that our method outperforms all the other methods by clear margins. This demonstrates that our model has promising transferability.

\subsection{Ablation Study}
\label{subsec:ablation study}

\begin{table}[t]
\small
\begin{center}
\adjustbox{max width=\linewidth}{
\setlength{\tabcolsep}{3.0pt}
\begin{tabular}{c|ccc|cc|cc|cc}
\toprule
\multirow{2}{*}{Method} & \multirow{2}{*}{CP} & \multirow{2}{*}{$\mathcal{L}_{\mathrm{RC}}$} & \multirow{2}{*}{$\mathcal{L}_{\mathrm{Q}}$}  & \multicolumn{2}{c|}{Market-1501} & \multicolumn{2}{c|}{MSMT17} & \multicolumn{2}{c}{CUHK03} \\ \cline{5-10}
&  &   &   & R1  & mAP   & R1  & mAP  & R1  & mAP   \\  \hline
MoCo~\cite{MoCo} & \ding{55}  & \ding{55}  & \ding{55} & 10.5 & 2.6 & 0.5 & 0.2 & 0.3 & 0.7 \\  \hline
\multirow{4}{*}{Ours} & \ding{51} &  \ding{55}  & \ding{55}  & 54.5  & 31.9  & 17.8  & 3.7 & 9.5  & 9.8            \\
& \ding{51} &  \ding{51}  & \ding{55}   & 81.5 & 58.7   & 41.4 & 18.2  & 24.9 & 26.0 \\
& \ding{51} &  \ding{55}  & \ding{51}   & 55.7 & 32.2   & 19.2 & 6.9  & 11.1 & 11.7 \\
& \ding{51} &  \ding{51}  & \ding{51}   & \textbf{85.1} & \textbf{65.1}   & \textbf{45.7} & \textbf{21.2}  & \textbf{26.1} & \textbf{27.4} \\
\bottomrule
\end{tabular}
}
\end{center}
\caption{Ablation study of the components under the DG settings. CP: Cross-frame positive Pairs; $\mathcal{L}_{\mathrm{RC}}$: reliability-guided contrastive loss; $\mathcal{L}_{\mathrm{Q}}$: memory-based contrastive loss.}
\label{table:ablation study of each components.}
\end{table}

\paragraph{The effectiveness of each component}. Table~\ref{table:ablation study of each components.} reports the ablation study of each component. There are three main components in our method, \ie, CP: cross-frame positive pairs; $\mathcal{L}_{\mathrm{RC}}$: reliability-guided contrastive loss in Eq.~\ref{eq:reliability-guided contrastive loss}; $\mathcal{L}_{\mathrm{Q}}$: memory-based contrastive loss in Eq.~\ref{eq:memory-based contrastive loss}.
We regard MoCo~\cite{MoCo} as a comparable baseline. We can make several observations.
Firstly, even without using $\mathcal{L}_{\mathrm{RC}}$ (set $\gamma$=0 in Eq.~\ref{eq:reliability-guided contrastive loss}) and $\mathcal{L}_{\mathrm{Q}}$, our framework outperforms MoCo by a large margin (\eg, +44.0\% Rank-1 in Market-1501). This confirms the effectiveness of our framework in learning identity discrimination by constructing positive pairs from inter-frame images instead of considering two augmented two augmented views of an image as a positive pair. Secondly, when $\mathcal{L}_{\mathrm{RC}}$ is applied, the performance is significantly improved, \eg, from 54.5\% to 81.5\% (+27.0\%) Rank-1 in Market-1501. This demonstrates that our $\mathcal{L}_{\mathrm{RC}}$ can effectively suppress the adverse impact of noisy positive pairs, resulting in more identity-discriminative representations. Thirdly, $\mathcal{L}_{\mathrm{Q}}$ helps to improve the discrimination of learned representations. For example, when adding $\mathcal{L}_{\mathrm{Q}}$ to $\mathcal{L}_{\mathrm{RC}}$, the Rank-1 on Market-1501 is improved from 81.5\% to 85.1\%. In conclusion, constructing positive pairs from adjacent frames is the basis of our method. $\mathcal{L}_{\mathrm{RC}}$ is much more important than $\mathcal{L}_{\mathrm{Q}}$. Combining all proposed components achieves the best performance.

\begin{table}[t]
\small
\begin{center}
\adjustbox{max width=\linewidth}{
\setlength{\tabcolsep}{3.0pt}
\begin{tabular}{c|cc|cc|cc}
\toprule
\multirow{2}{*}{Loss function}& \multicolumn{2}{c|}{Market-1501} & \multicolumn{2}{c|}{MSMT17} & \multicolumn{2}{c}{CUHK03} \\ \cline{2-7}
& R1  & mAP   & R1  & mAP  & R1  & mAP   \\  \hline
Focal loss~\cite{focal_loss} & 26.7 & 10.1 & 4.6 & 1.6 & 0.7 & 1.2 \\
$\mathcal{L}_{\mathrm{RC}}$  & \textbf{85.1}   & \textbf{65.1} & \textbf{45.7}   & \textbf{21.2} & \textbf{26.1}  & \textbf{27.4} \\
\bottomrule
\end{tabular}
}
\end{center}
\caption{Comparisons with focal loss~\cite{focal_loss} under the DG settings.}
\label{table:comparison with focal loss.}
\end{table}

\begin{figure}[t]
    \centering
    \includegraphics[width=0.95\linewidth, height=0.62\linewidth]{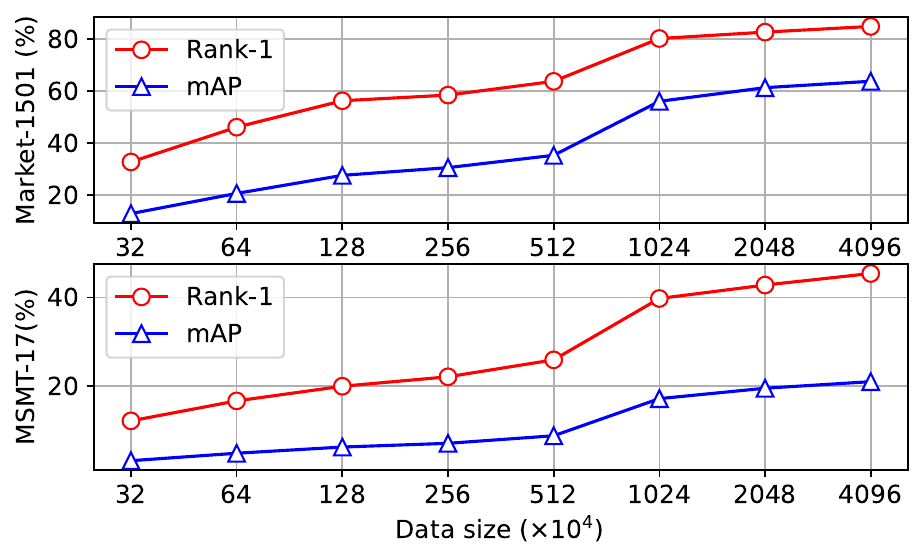}
    \caption{Performance \vs  data scale. The performance keeps improving as the training data increases and does not saturate.}
    \label{fig:performance_vs_data_scale}
\end{figure}

\paragraph{Comparison with focal loss~\cite{focal_loss}}. We conduct comparative experiments with focal loss to verify the effectiveness of our $\mathcal{L}_{\mathrm{RC}}$. Specifically, we only replace $\mathcal{L}_{\mathrm{RC}}$ with focal loss, and keep other settings unchanged.
As shown in Table~\ref{table:comparison with focal loss.}, $\mathcal{L}_{\mathrm{RC}}$ outperforms focal loss by a considerable margin. This supports the notion that it is error-prone to assign large training weight to hard samples in the presence of noisy labels since the label of the hard sample is likely to be incorrect. Instead, it is more appropriate to guide the model to learn from reliable samples.

\paragraph{Performance \vs Data size}. Here, we study the effect of the training data size for our method. Under the DG ReID settings, we vary the training data size from $32\times10^{4}$ to $2048\times10^{4}$ to see how the performance changes. The backbone is ResNet50. We report the results in Figure~\ref{fig:performance_vs_data_scale}. As we can see, with the increase of the data scale, the performance on Market-1501 and MSMT17 consistently keeps improving. Even at the large scale of $4096\times10^4$ unlabeled images, the performance does not saturate. There is still a clear upward trend in Rank-1 of MSMT17. This shows that our method has excellent scalability to large-scale unlabeled data, which has the potential to achieve better results as the training data continues to grow.

\begin{table}[t]
\small
\begin{center}
\adjustbox{max width=\linewidth}{
\setlength{\tabcolsep}{3.0pt}
\begin{tabular}{c|c|c|c|c|c}
\toprule
 $\delta_{\mathrm{max}}$ & 0.5s & 1.0s & 2.0s & 4.0s  & 8.0s \\ \hline
Market & 73.7/48.4 & 81.4/58.4 & 83.8/63.2  & \textbf{85.1/65.1}  & 84.5/64.4 \\
MSMT17 & 34.6/13.8 & 41.2/17.9 & 43.5/19.0  & \textbf{45.7/21.2}  & 44.8/20.1 \\
\bottomrule
\end{tabular}
}
\end{center}
\caption{Impact of the maximum time interval $\delta_{\mathrm{max}}$. \textit{Rank-1/mAP}}
\label{table:ablation study of maximum time interval.}
\end{table}

\paragraph{Time interval between two frame}. In this part, we investigate the impact of the maximum time interval $\delta_{\mathrm{max}}$. As in Table~\ref{table:ablation study of maximum time interval.}, the choice of the maximum time interval involves trade-offs. When $\delta_{\mathrm{max}}$ is small, the performance increases as $\delta_{\mathrm{max}}$ increases because richer contrastive information is provided. When $\delta_{\mathrm{max}}$ is too large, the performance degrades because the mined positive pairs are error-prone, misleading the representation learning. Finally, we set $\delta_{\mathrm{max}}=4.0s$ in our experiments.

\begin{figure}[t]
    \centering
    \includegraphics[width=\linewidth, height=0.49\linewidth]{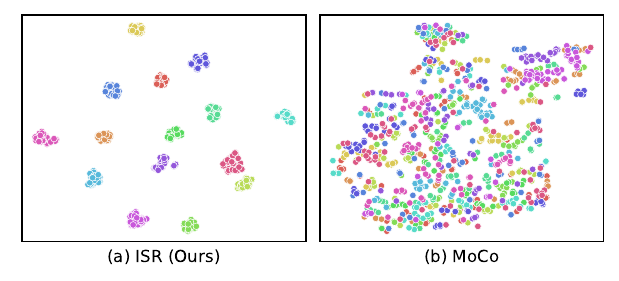}
    \caption{The t-SNE~\cite{tsne} visualization of the representation. Our method learns identity discrimination, while MoCo~\cite{MoCo} does not.}
    \label{fig:tsne}
\end{figure}

\paragraph{Feature distribution}. Here, we investigate the feature distribution to validate the capability of our method in learning identity discrimination. We collect samples of 15 identities that are randomly selected from the gallery of the Market-1501~\cite{market}, and visualize their features via t-SNE~\cite{tsne} in Figure~\ref{fig:tsne}. MoCo~\cite{MoCo} is employed as a comparison.  It can be seen that the features extracted by our model are closely clustered for all images of a given identity, while the features extracted by MoCo are dispersed. This observation verifies that MoCo can only learn instance-discriminative representations, which is determined by the way positive pairs are constructed. More importantly, this observation provides compelling evidence that our method effectively learns identity-discriminative representations.

\begin{figure}[t]
    \centering
    \includegraphics[width=\linewidth, height=0.45\linewidth]{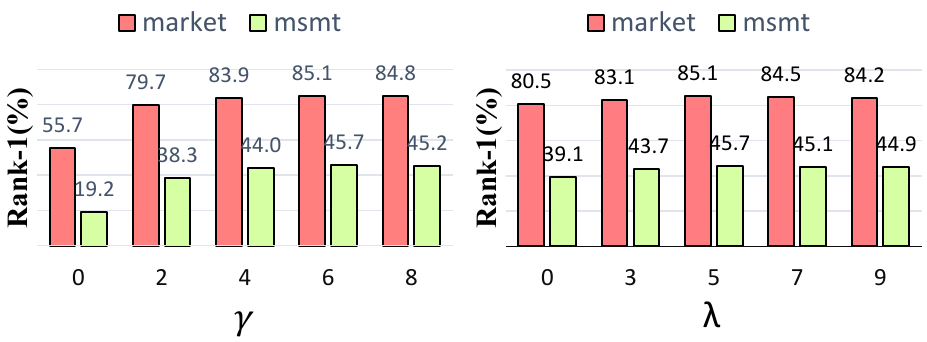}
    \caption{Analysis of hyper-parameters $\gamma$ in Eq.~\ref{eq:reliability-guided contrastive loss} and $\lambda$ in Eq.~\ref{eq:final loss}.}
    \label{fig:hyper_para}
\end{figure}

\paragraph{Impact of hyper-parameters}. Here, we show how $\gamma$ in Eq.~\ref{eq:reliability-guided contrastive loss} and $\lambda$ in Eq.~\ref{eq:final loss} affect the performance. Experiments are conducted on Market-1501 and MSMT17 under the DG ReID settings.  The backbone is ResNet-50. We report the results in Figure~\ref{fig:hyper_para}. 
We can see that $\gamma=2,4,6,8$ improves $\gamma=0$ a lot and $\lambda=3,5,7,9$ outperforms $\lambda=0$. This again shows the effectiveness of the proposed reliability-guided contrastive loss and memory-based contrastive loss. Our method is robust to $\lambda$ and $\gamma$.
Finally, the optimal values of $\gamma$ and $\lambda$ are 6 and 5, respectively.

\section{Conclusions}
\label{sec:conclusions}
This paper proposes an identity-seeking self-supervised representation learning method for learning a generalizable ReID representation from large-scale unlabeled videos. Unlike conventional contrastive learning methods that learn instance discrimination, our method learns identity discrimination by enforcing inter-frame instances with the same identity to have similar representations. The learned representation generalizes well to unseen domains without fine-tuning. In addition, the learned representation exhibits promising adaptability. We hope our method will provide new insights and attract more interest in large-scale unsupervised learning for domain-generalizable ReID.

\paragraph{Acknowledgements.} This work was supported by the National Key Research and Development Program of China in the 14th Five-Year (Nos. 2021YFFO602103 and 2021YFF0602102).

{\small
\bibliographystyle{ieee_fullname}
\bibliography{egbib}

\begin{thebibliography}{10}\itemsep=-1pt

\bibitem{bai2021hierarchical}
Yan Bai, Ce Wang, Yihang Lou, Jun Liu, and Ling-Yu Duan.
\newblock Hierarchical connectivity-centered clustering for unsupervised domain
  adaptation on person re-identification.
\newblock {\em IEEE TIP}, pages 6715--6729, 2021.

\bibitem{chen2019mixed}
Binghui Chen, Weihong Deng, and Jiani Hu.
\newblock Mixed high-order attention network for person re-identification.
\newblock In {\em ICCV}, pages 371--381, 2019.

\bibitem{MHN}
Binghui Chen, Weihong Deng, and Jiani Hu.
\newblock Mixed high-order attention network for person re-identification.
\newblock In {\em ICCV}, pages 371--381, 2019.

\bibitem{chen2019self}
Guangyi Chen, Chunze Lin, Liangliang Ren, Jiwen Lu, and Jie Zhou.
\newblock Self-critical attention learning for person re-identification.
\newblock In {\em ICCV}, pages 9637--9646, 2019.

\bibitem{DDAN}
Peixian Chen, Pingyang Dai, Jianzhuang Liu, Feng Zheng, Mingliang Xu, Qi Tian,
  and Rongrong Ji.
\newblock Dual distribution alignment network for generalizable person
  re-identification.
\newblock In {\em AAAI}, pages 1054--1062, 2021.

\bibitem{ABD-Net}
Tianlong Chen, Shaojin Ding, Jingyi Xie, Ye Yuan, Wuyang Chen, Yang Yang, Zhou
  Ren, and Zhangyang Wang.
\newblock Abd-net: Attentive but diverse person re-identification.
\newblock In {\em ICCV}, pages 8351--8361, 2019.

\bibitem{ABDNet}
Tianlong Chen, Shaojin Ding, Jingyi Xie, Ye Yuan, Wuyang Chen, Yang Yang, Zhou
  Ren, and Zhangyang Wang.
\newblock Abd-net: Attentive but diverse person re-identification.
\newblock In {\em ICCV}, pages 8351--8361, 2019.

\bibitem{SimCLR}
Ting Chen, Simon Kornblith, Mohammad Norouzi, and Geoffrey Hinton.
\newblock A simple framework for contrastive learning of visual
  representations.
\newblock In {\em ICML}, pages 1597--1607, 2020.

\bibitem{tripletloss}
Weihua Chen, Xiaotang Chen, Jianguo Zhang, and Kaiqi Huang.
\newblock Beyond triplet loss: a deep quadruplet network for person
  re-identification.
\newblock In {\em CVPR}, pages 403--412, 2017.

\bibitem{simsiam}
Xinlei Chen and Kaiming He.
\newblock Exploring simple siamese representation learning.
\newblock In {\em CVPR}, pages 15750--15758, 2021.

\bibitem{MetaBIN}
Seokeon Choi, Taekyung Kim, Minki Jeong, Hyoungseob Park, and Changick Kim.
\newblock Meta batch-instance normalization for generalizable person
  re-identification.
\newblock In {\em CVPR}, pages 3425--3435, 2021.

\bibitem{RaMoE}
Yongxing Dai, Xiaotong Li, Jun Liu, Zekun Tong, and Ling-Yu Duan.
\newblock Generalizable person re-identification with relevance-aware mixture
  of experts.
\newblock In {\em CVPR}, pages 16145--16154, 2021.

\bibitem{BDB}
Zuozhuo Dai, Mingqiang Chen, Xiaodong Gu, Siyu Zhu, and Ping Tan.
\newblock Batch dropblock network for person re-identification and beyond.
\newblock In {\em ICCV}, pages 3691--3701, 2019.

\bibitem{UAL}
Zhaopeng Dou, Zhongdao Wang, Weihua Chen, Yali Li, and Shengjin Wang.
\newblock Reliability-aware prediction via uncertainty learning for person
  image retrieval.
\newblock In {\em ECCV}, pages 588--605, 2022.

\bibitem{feng2021complementary}
Hao Feng, Minghao Chen, Jinming Hu, Dong Shen, Haifeng Liu, and Deng Cai.
\newblock Complementary pseudo labels for unsupervised domain adaptation on
  person re-identification.
\newblock {\em IEEE TIP}, pages 2898--2907, 2021.

\bibitem{LUP}
Dengpan Fu, Dongdong Chen, Jianmin Bao, Hao Yang, Lu Yuan, Lei Zhang, Houqiang
  Li, and Dong Chen.
\newblock Unsupervised pre-training for person re-identification.
\newblock In {\em CVPR}, pages 14750--14759, 2021.

\bibitem{LUP-NL}
Dengpan Fu, Dongdong Chen, Hao Yang, Jianmin Bao, Lu Yuan, Lei Zhang, Houqiang
  Li, Fang Wen, and Dong Chen.
\newblock Large-scale pre-training for person re-identification with noisy
  labels.
\newblock In {\em CVPR}, pages 2476--2486, 2022.

\bibitem{IIA}
Dengpan Fu, Bo Xin, Jingdong Wang, Dongdong Chen, Jianmin Bao, Gang Hua, and
  Houqiang Li.
\newblock Improving person re-identification with iterative impression
  aggregation.
\newblock {\em IEEE TIP}, pages 9559--9571, 2020.

\bibitem{fu2019self}
Yang Fu, Yunchao Wei, Guanshuo Wang, Yuqian Zhou, Honghui Shi, and Thomas~S
  Huang.
\newblock Self-similarity grouping: A simple unsupervised cross domain
  adaptation approach for person re-identification.
\newblock In {\em ICCV}, pages 6112--6121, 2019.

\bibitem{VIPeR}
Douglas Gray and Hai Tao.
\newblock Viewpoint invariant pedestrian recognition with an ensemble of
  localized features.
\newblock In {\em ECCV}, pages 262--275, 2008.

\bibitem{BYOL}
Jean-Bastien Grill, Florian Strub, Florent Altch{\'e}, Corentin Tallec, Pierre
  Richemond, Elena Buchatskaya, Carl Doersch, Bernardo Avila~Pires, Zhaohan
  Guo, Mohammad Gheshlaghi~Azar, et~al.
\newblock Bootstrap your own latent-a new approach to self-supervised learning.
\newblock {\em NeurIPS}, pages 21271--21284, 2020.

\bibitem{gu2022autoloss}
Hongyang Gu, Jianmin Li, Guangyuan Fu, Chifong Wong, Xinghao Chen, and Jun Zhu.
\newblock Autoloss-gms: Searching generalized margin-based softmax loss
  function for person re-identification.
\newblock In {\em CVPR}, pages 4744--4753, 2022.

\bibitem{guo2019beyond}
Jianyuan Guo, Yuhui Yuan, Lang Huang, Chao Zhang, Jin-Ge Yao, and Kai Han.
\newblock Beyond human parts: Dual part-aligned representations for person
  re-identification.
\newblock In {\em ICCV}, pages 3642--3651, 2019.

\bibitem{MAE}
Kaiming He, Xinlei Chen, Saining Xie, Yanghao Li, Piotr Doll{\'a}r, and Ross
  Girshick.
\newblock Masked autoencoders are scalable vision learners.
\newblock In {\em CVPR}, pages 16000--16009, 2022.

\bibitem{MoCo}
Kaiming He, Haoqi Fan, Yuxin Wu, Saining Xie, and Ross Girshick.
\newblock Momentum contrast for unsupervised visual representation learning.
\newblock In {\em CVPR}, pages 9729--9738, 2020.

\bibitem{resnet}
Kaiming He, Xiangyu Zhang, Shaoqing Ren, and Jian Sun.
\newblock Deep residual learning for image recognition.
\newblock In {\em CVPR}, pages 770--778, 2016.

\bibitem{GASM}
Lingxiao He and Wu Liu.
\newblock Guided saliency feature learning for person re-identification in
  crowded scenes.
\newblock In {\em ECCV}, pages 357--373, 2020.

\bibitem{Transreid}
Shuting He, Hao Luo, Pichao Wang, Fan Wang, Hao Li, and Wei Jiang.
\newblock Transreid: Transformer-based object re-identification.
\newblock In {\em ICCV}, pages 15013--15022, 2021.

\bibitem{hermans2017defense}
Alexander Hermans, Lucas Beyer, and Bastian Leibe.
\newblock In defense of the triplet loss for person re-identification.
\newblock {\em arXiv preprint arXiv:1703.07737}, 2017.

\bibitem{PRID}
Martin Hirzer, Csaba Beleznai, Peter~M Roth, and Horst Bischof.
\newblock Person re-identification by descriptive and discriminative
  classification.
\newblock In {\em Scandinavian conference on Image analysis}, pages 91--102,
  2011.

\bibitem{RFC}
Ruibing Hou, Bingpeng Ma, Hong Chang, Xinqian Gu, Shiguang Shan, and Xilin
  Chen.
\newblock Feature completion for occluded person re-identification.
\newblock {\em IEEE TPAMI}, 2021.

\bibitem{TrackContrast}
Weiquan Huang, Yan Bai, Qiuyu Ren, Xinbo Zhao, Ming Feng, and Yin Wang.
\newblock Large-scale unsupervised person re-identification with contrastive
  learning.
\newblock {\em arXiv preprint arXiv:2105.07914}, 2021.

\bibitem{jia2019frustratingly}
Jieru Jia, Qiuqi Ruan, and Timothy~M Hospedales.
\newblock Frustratingly easy person re-identification: Generalizing person
  re-id in practice.
\newblock {\em arXiv preprint arXiv:1905.03422}, 2019.

\bibitem{MoS}
Mengxi Jia, Xinhua Cheng, Yunpeng Zhai, Shijian Lu, Siwei Ma, Yonghong Tian,
  and Jian Zhang.
\newblock Matching on sets: Conquer occluded person re-identification without
  alignment.
\newblock In {\em AAAI}, pages 1673--1681, 2021.

\bibitem{DTIN-Net}
Bingliang Jiao, Lingqiao Liu, Liying Gao, Guosheng Lin, Lu Yang, Shizhou Zhang,
  Peng Wang, and Yanning Zhang.
\newblock Dynamically transformed instance normalization network for
  generalizable person re-identification.
\newblock In {\em ECCV}, pages 285--301, 2022.

\bibitem{SNR}
Xin Jin, Cuiling Lan, Wenjun Zeng, Zhibo Chen, and Li Zhang.
\newblock Style normalization and restitution for generalizable person
  re-identification.
\newblock In {\em CVPR}, pages 3143--3152, 2020.

\bibitem{hungarian}
Harold~W Kuhn.
\newblock The hungarian method for the assignment problem.
\newblock {\em Naval research logistics quarterly}, pages 83--97, 1955.

\bibitem{MLDG}
Da Li, Yongxin Yang, Yi-Zhe Song, and Timothy Hospedales.
\newblock Learning to generalize: Meta-learning for domain generalization.
\newblock In {\em AAAI}, 2018.

\bibitem{li2021combined}
Hanjun Li, Gaojie Wu, and Wei-Shi Zheng.
\newblock Combined depth space based architecture search for person
  re-identification.
\newblock In {\em CVPR}, pages 6729--6738, 2021.

\bibitem{CUHK02}
Wei Li and Xiaogang Wang.
\newblock Locally aligned feature transforms across views.
\newblock In {\em CVPR}, pages 3594--3601, 2013.

\bibitem{CUHK03}
Wei Li, Rui Zhao, Tong Xiao, and Xiaogang Wang.
\newblock Deepreid: Deep filter pairing neural network for person
  re-identification.
\newblock In {\em CVPR}, pages 152--159, 2014.

\bibitem{li2021diverse}
Yulin Li, Jianfeng He, Tianzhu Zhang, Xiang Liu, Yongdong Zhang, and Feng Wu.
\newblock Diverse part discovery: Occluded person re-identification with
  part-aware transformer.
\newblock In {\em CVPR}, pages 2898--2907, 2021.

\bibitem{focal_loss}
Tsung-Yi Lin, Priya Goyal, Ross Girshick, Kaiming He, and Piotr Doll{\'a}r.
\newblock Focal loss for dense object detection.
\newblock In {\em CVPR}, pages 2980--2988, 2017.

\bibitem{swin-transformer}
Ze Liu, Yutong Lin, Yue Cao, Han Hu, Yixuan Wei, Zheng Zhang, Stephen Lin, and
  Baining Guo.
\newblock Swin transformer: Hierarchical vision transformer using shifted
  windows.
\newblock In {\em ICCV}, pages 10012--10022, 2021.

\bibitem{CosineAnnealing}
Ilya Loshchilov and Frank Hutter.
\newblock Sgdr: Stochastic gradient descent with warm restarts.
\newblock {\em arXiv preprint arXiv:1608.03983}, 2016.

\bibitem{AdamW}
Ilya Loshchilov and Frank Hutter.
\newblock Decoupled weight decay regularization.
\newblock {\em arXiv preprint arXiv:1711.05101}, 2017.

\bibitem{grid}
Chen~Change Loy, Tao Xiang, and Shaogang Gong.
\newblock Time-delayed correlation analysis for multi-camera activity
  understanding.
\newblock {\em IJCV}, pages 106--129, 2010.

\bibitem{BOT}
Hao Luo, Wei Jiang, Youzhi Gu, Fuxu Liu, Xingyu Liao, Shenqi Lai, and Jianyang
  Gu.
\newblock A strong baseline and batch normalization neck for deep person
  re-identification.
\newblock {\em IEEE TMM}, pages 2597--2609, 2019.

\bibitem{MDA}
Hao Ni, Jingkuan Song, Xiaopeng Luo, Feng Zheng, Wen Li, and Heng~Tao Shen.
\newblock Meta distribution alignment for generalizable person
  re-identification.
\newblock In {\em CVPR}, pages 2487--2496, 2022.

\bibitem{IBN}
Xingang Pan, Ping Luo, Jianping Shi, and Xiaoou Tang.
\newblock Two at once: Enhancing learning and generalization capacities via
  ibn-net.
\newblock In {\em ECCV}, pages 464--479, 2018.

\bibitem{GCP}
Hyunjong Park and Bumsub Ham.
\newblock Relation network for person re-identification.
\newblock In {\em AAAI}, pages 11839--11847, 2020.

\bibitem{PPA}
Siyuan Qiao, Chenxi Liu, Wei Shen, and Alan~L Yuille.
\newblock Few-shot image recognition by predicting parameters from activations.
\newblock In {\em CVPR}, pages 7229--7238, 2018.

\bibitem{rami2022online}
Hamza Rami, Matthieu Ospici, and St{\'e}phane Lathuili{\`e}re.
\newblock Online unsupervised domain adaptation for person re-identification.
\newblock In {\em CVPR}, pages 3830--3839, 2022.

\bibitem{duke}
Ergys Ristani, Francesco Solera, Roger Zou, Rita Cucchiara, and Carlo Tomasi.
\newblock Performance measures and a data set for multi-target, multi-camera
  tracking.
\newblock In {\em ECCV}, pages 17--35, 2016.

\bibitem{CrossGrad}
Shiv Shankar, Vihari Piratla, Soumen Chakrabarti, Siddhartha Chaudhuri, Preethi
  Jyothi, and Sunita Sarawagi.
\newblock Generalizing across domains via cross-gradient training.
\newblock {\em arXiv preprint arXiv:1804.10745}, 2018.

\bibitem{ESNET}
Dong Shen, Shuai Zhao, Jinming Hu, Hao Feng, Deng Cai, and Xiaofei He.
\newblock Es-net: Erasing salient parts to learn more in re-identification.
\newblock {\em IEEE TIP}, pages 1676--1686, 2020.

\bibitem{DIMN}
Jifei Song, Yongxin Yang, Yi-Zhe Song, Tao Xiang, and Timothy~M Hospedales.
\newblock Generalizable person re-identification by domain-invariant mapping
  network.
\newblock In {\em CVPR}, pages 719--728, 2019.

\bibitem{PersonX}
Xiaoxiao Sun and Liang Zheng.
\newblock Dissecting person re-identification from the viewpoint of viewpoint.
\newblock In {\em CVPR}, pages 608--617, 2019.

\bibitem{circleloss}
Yifan Sun, Changmao Cheng, Yuhan Zhang, Chi Zhang, Liang Zheng, Zhongdao Wang,
  and Yichen Wei.
\newblock Circle loss: A unified perspective of pair similarity optimization.
\newblock In {\em CVPR}, pages 6398--6407, 2020.

\bibitem{PCB}
Yifan Sun, Liang Zheng, Yi Yang, Qi Tian, and Shengjin Wang.
\newblock Beyond part models: Person retrieval with refined part pooling (and a
  strong convolutional baseline).
\newblock In {\em ECCV}, pages 480--496, 2018.

\bibitem{IN}
Dmitry Ulyanov, Andrea Vedaldi, and Victor Lempitsky.
\newblock Improved texture networks: Maximizing quality and diversity in
  feed-forward stylization and texture synthesis.
\newblock In {\em CVPR}, pages 6924--6932, 2017.

\bibitem{tsne}
Laurens Van Der~Maaten.
\newblock Accelerating t-sne using tree-based algorithms.
\newblock {\em The journal of machine learning research}, pages 3221--3245,
  2014.

\bibitem{MGN}
Guanshuo Wang, Yufeng Yuan, Xiong Chen, Jiwei Li, and Xi Zhou.
\newblock Learning discriminative features with multiple granularities for
  person re-identification.
\newblock In {\em ACM MM}, pages 274--282, 2018.

\bibitem{iLIDs}
Taiqing Wang, Shaogang Gong, Xiatian Zhu, and Shengjin Wang.
\newblock Person re-identification by video ranking.
\newblock In {\em ECCV}, pages 688--703, 2014.

\bibitem{wang2022pose}
Tao Wang, Hong Liu, Pinhao Song, Tianyu Guo, and Wei Shi.
\newblock Pose-guided feature disentangling for occluded person
  re-identification based on transformer.
\newblock In {\em AAAI}, pages 2540--2549, 2022.

\bibitem{Domain-mix}
Wenhao Wang, Shengcai Liao, Fang Zhao, Kangkang Cui, and Ling Shao.
\newblock Domainmix: Learning generalizable person re-identification without
  human annotations.
\newblock In {\em BMVC}, 2021.

\bibitem{wang2022attentive}
Wenhao Wang, Fang Zhao, Shengcai Liao, and Ling Shao.
\newblock Attentive waveblock: complementarity-enhanced mutual networks for
  unsupervised domain adaptation in person re-identification and beyond.
\newblock {\em IEEE TIP}, pages 1532--1544, 2022.

\bibitem{XBM}
Xun Wang, Haozhi Zhang, Weilin Huang, and Matthew~R Scott.
\newblock Cross-batch memory for embedding learning.
\newblock In {\em CVPR}, pages 6388--6397, 2020.

\bibitem{ClonedPerson}
Yanan Wang, Xuezhi Liang, and Shengcai Liao.
\newblock Cloning outfits from real-world images to 3d characters for
  generalizable person re-identification.
\newblock In {\em CVPR}, pages 4900--4909, 2022.

\bibitem{RandPerson}
Yanan Wang, Shengcai Liao, and Ling Shao.
\newblock Surpassing real-world source training data: Random 3d characters for
  generalizable person re-identification.
\newblock In {\em ACM MM}, pages 3422--3430, 2020.

\bibitem{CycAs}
Zhongdao Wang, Zhaopeng Dou, Jingwei Zhang, Liang Zhen, Yifan Sun, Yali Li, and
  Shengjin Wang.
\newblock Generalizable re-identification from videos with cycle association.
\newblock {\em arXiv preprint arXiv:2211.03663}, 2022.

\bibitem{cycaseccv}
Zhongdao Wang, Jingwei Zhang, Liang Zheng, Yixuan Liu, Yifan Sun, Yali Li, and
  Shengjin Wang.
\newblock Cycas: Self-supervised cycle association for learning re-identifiable
  descriptions.
\newblock In {\em ECCV}, 2020.

\bibitem{JDE}
Zhongdao Wang, Liang Zheng, Yixuan Liu, Yali Li, and Shengjin Wang.
\newblock Towards real-time multi-object tracking.
\newblock In {\em ECCV}, pages 107--122, 2020.

\bibitem{MSMT17}
Longhui Wei, Shiliang Zhang, Wen Gao, and Qi Tian.
\newblock Person transfer gan to bridge domain gap for person
  re-identification.
\newblock In {\em CVPR}, pages 79--88, 2018.

\bibitem{CUHK-SYSU}
Tong Xiao, Shuang Li, Bochao Wang, Liang Lin, and Xiaogang Wang.
\newblock End-to-end deep learning for person search.
\newblock {\em arXiv preprint arXiv:1604.01850}, 2016.

\bibitem{META}
Boqiang Xu, Jian Liang, Lingxiao He, and Zhenan Sun.
\newblock Mimic embedding via adaptive aggregation: learning generalizable
  person re-identification.
\newblock In {\em ECCV}, pages 372--388, 2022.

\bibitem{ACL}
Pengyi Zhang, Huanzhang Dou, Yunlong Yu, and Xi Li.
\newblock Adaptive cross-domain learning for generalizable person
  re-identification.
\newblock In {\em ECCV}, pages 215--232, 2022.

\bibitem{UnrealPerson}
Tianyu Zhang, Lingxi Xie, Longhui Wei, Zijie Zhuang, Yongfei Zhang, Bo Li, and
  Qi Tian.
\newblock Unrealperson: An adaptive pipeline towards costless person
  re-identification.
\newblock In {\em CVPR}, pages 11506--11515, 2021.

\bibitem{DIR_ReID}
Yi-Fan Zhang, Zhang Zhang, Da Li, Zhen Jia, Liang Wang, and Tieniu Tan.
\newblock Learning domain invariant representations for generalizable person
  re-identification.
\newblock {\em IEEE TIP}, 2022.

\bibitem{DSA}
Zhizheng Zhang, Cuiling Lan, Wenjun Zeng, and Zhibo Chen.
\newblock Densely semantically aligned person re-identification.
\newblock In {\em CVPR}, pages 667--676, 2019.

\bibitem{M3L}
Yuyang Zhao, Zhun Zhong, Fengxiang Yang, Zhiming Luo, Yaojin Lin, Shaozi Li,
  and Nicu Sebe.
\newblock Learning to generalize unseen domains via memory-based multi-source
  meta-learning for person re-identification.
\newblock In {\em CVPR}, pages 6277--6286, 2021.

\bibitem{market}
Liang Zheng, Liyue Shen, Lu Tian, Shengjin Wang, Jingdong Wang, and Qi Tian.
\newblock Scalable person re-identification: A benchmark.
\newblock In {\em ICCV}, pages 1116--1124, 2015.

\bibitem{ReRank}
Zhun Zhong, Liang Zheng, Donglin Cao, and Shaozi Li.
\newblock Re-ranking person re-identification with k-reciprocal encoding.
\newblock In {\em CVPR}, pages 1318--1327, 2017.

\bibitem{zhong2019invariance}
Zhun Zhong, Liang Zheng, Zhiming Luo, Shaozi Li, and Yi Yang.
\newblock Invariance matters: Exemplar memory for domain adaptive person
  re-identification.
\newblock In {\em CVPR}, pages 598--607, 2019.

\bibitem{OSNet}
Kaiyang Zhou, Yongxin Yang, Andrea Cavallaro, and Tao Xiang.
\newblock Omni-scale feature learning for person re-identification.
\newblock In {\em ICCV}, pages 3702--3712, 2019.

\bibitem{ISP}
Kuan Zhu, Haiyun Guo, Zhiwei Liu, Ming Tang, and Jinqiao Wang.
\newblock Identity-guided human semantic parsing for person re-identification.
\newblock In {\em ECCV}, pages 346--363, 2020.

\end{thebibliography}
}

\end{document}